\documentclass[10pt,twocolumn,letterpaper]{article}

\usepackage{iccv,times,graphicx,tabulary,multirow,subfig,xspace,xcolor}
\usepackage{amsthm,amssymb,fixmath,mathtools,nicefrac}
\usepackage[pagebackref=true,breaklinks=true,letterpaper=true,colorlinks,
  citecolor=citecolor,bookmarks=false]{hyperref}
\definecolor{citecolor}{RGB}{34,139,34}

\begin{document}
\title{Towards Good Practices for Instance Segmentation}
\track{COCO Instance Segmentation}

\author{
Dongdong Yu, Zehuan Yuan, Jinlai Liu, Kun Yuan, Changhu Wang \\
ByteDance AI Lab, China \\
}

\maketitle

\begin{abstract}
Instance Segmentation is an interesting yet challenging task in computer vision. In this paper, we conduct a series of refinements with the Hybrid Task Cascade (HTC) Network, and empirically evaluate their impact on the final model performance through ablation studies. By taking all the refinements, we achieve 0.47 on the COCO test-dev dataset and 0.47 on the COCO test-challenge dataset.
\end{abstract}

\section{Introduction}
Instance segmentation is an interesting yet challenging task in computer vision. The goal is to label each pixel into foreground or backbone for each object instance detection bounding-box.
It is important for many applications, such as autonomous driving, virtual reality, human-computer interaction and activity recognition.

Recently, the problem of instance segmentation has been greatly improved by the involvement of deep convolutional neural networks~\cite{he2016deep,chen2019mmdetection,chen2019hybrid,oksuz2019imbalance}. Existing approaches are mostly follows the Mask-RCNN method~\cite{he2017mask}, which extends Faster R-CNN by adding a branch for predicting an object mask in parallel with the existing branch for bounding box recognition. For example, Hybrid Task Cascade (HTC) Network~\cite{chen2019hybrid} interweaves the box detection task and instance segmentation task for a joint multi-stage refinement processing and devils spatial context information to help distinguishing hard foreground from cluttered background, which ranks 1st in the COCO 2018 Challenge Object Detection Task. There are many tricks to improve the performance of instance segmentation, such as, strong backbone,  balanced learning strategy, effective sampling strategy, multi-scale training, multi-scale testing and model ensemble.

In this paper, we follow the Hybrid Task Cascade Network pipeline and conduct a series of refinements(new backbone, nms strategy and re-compute the mask score) based on the Hybrid Task Cascade Network  and evaluate their impact on the final model performance through ablation studies. Finally, we achieve 0.47 on the COCO test-dev dataset and 0.47 on the COCO test- challenge dataset.

\section{Method}
To handle the instance segmentation, we follow the Hybrid Task Cascade Network pipeline to detect each instance in the image and then classify the foreground and backbone pixels for each detected instance. 

The HTC network can adopts the ResNet, ResNext or SENet as the backbone of the feature encoder. In our work, we propose a new backbone, named RefineNet~\cite{yu2019towards}, which can well handle the scale variant cases. Different than most algorithms, we proposed a post-processing method to re-compute the mask score, instead of directly using the bounding-box score.

\section{Experiments}
 
\subsection{Datasets}
The training dataset only includes the COCO train2017 dataset~\cite{lin2014coco,caesar2018coco}, we do not use any other dataset. The final results are reported on the COCO test-dev dataset and the COCO test-challenge dataset.

\subsection{Results}
\subsubsection{Ablation Study}
In this subsection, we will step-wise decompose our model to reveal the effect of each component, including our new backbone, our re-compute mask score strategy and nms strategy. In the following experiments, we evaluate all comparisons on the COCO val2017 dataset.

\noindent{\textbf{Effect of Backbone}}~~In our paper, we modify the Residual bottleneck and propose a new backbone, named RefineNet, which is similar to ResNet, ResNext or SENet. Based on the HTC Network~\cite{chen2019mmdetection}, we implement the SENet154 HTC and SE-RefineNet154 HTC. As shown in Table~\ref{table:table1}, we implement the SENet154 in mmdetection, the AP performance is obvious lower than the author's implement, unfortunately. Based on our implement SENet154, we replace the bottleneck with our modified bottleneck, the AP is improved from 0.426 to 0.434.

\begin{table}[tb]
	\centering
	\caption{Results with different backbones on COCO val2017 dataset.}\label{table:table1}
	
	\resizebox{0.8\columnwidth}{!}{
	\begin{tabular}{cc}
		\hline
		Backbone  & AP \\
		\hline
		SENet154(Author's implement) & $\sim$0.446  \\
		SENet154(Our implement) & 0.426  \\
		SERefineNet154(Ours) & 0.434 \\
		\hline
	\end{tabular}
	}
\end{table}

\begin{table}[tb]
	\centering
	\caption{Results with re-compute score strategy on COCO val2017 dataset.}\label{table:table2}
	
	\resizebox{0.6\columnwidth}{!}{
	\begin{tabular}{cc}
		\hline
		Mask score  & AP \\
		\hline
		Bounding-box score & 0.434  \\
		 Re-computed score & 0.445  \\
		\hline
	\end{tabular}
	}
\end{table}

\begin{table}[tb]
	\centering
	\caption{Results with nms strategy on COCO val2017 dataset.}\label{table:table3}
	
	\resizebox{0.4\columnwidth}{!}{
	\begin{tabular}{cc}
		\hline
		Mask score  & AP \\
		\hline
		NMS & 0.445  \\
		 Soft-NMS & 0.447  \\
		\hline
	\end{tabular}
	}
\end{table}

\noindent{\textbf{Effect of Mask Score}}~~In our experiments, we find that it is not the best choice to use the bounding-box score as the mask score. We re-compute the score which takes the mask confidence score into account. As shown in Table~\ref{table:table2}, by using the re-compute score strategy, the AP is improved from 0.434 to 0.445.

\noindent{\textbf{Effect of NMS}}~~Besides, we also try the soft-nms strategy. As shown in Table~\ref{table:table3}, the AP is improved from 0.445 to 0.447.

\subsubsection{Development and Challenge Results}
In this subsection, we ensemble four RefineNet models, and take the multi-scale testing strategy and flip testing strategy for the instance segmentation. As shown in Table~\ref{table:table4}, the AP of test-dev is 0.473, and the AP of test-cha is 0.47. 

\begin{table}[tb]
	\centering
	\caption{Results of our model on COCO2017 test-dev and test-cha dataset.}\label{table:table4}
	
	\resizebox{0.9\columnwidth}{!}{
	\begin{tabular}{ccc}
		\hline
		BackBone & Development set & Challenge set \\
		\hline
		RefineNet & 0.473 & 0.47 \\
		\hline
	\end{tabular}
	}
\end{table}

\section{Conclusion}
In this paper, we conduct a series of refinements with the Hybrid Task Cascade (HTC) Network, and empirically evaluate their impact on the final model performance through ablation studies. By taking all the refinements, we achieve 0.47 on the COCO test-dev dataset and 0.47 on the COCO test-challenge dataset. 

{\small \bibliographystyle{ieee_fullname} \bibliography{egbib}}

\end{document}